\documentclass[letterpaper]{article} %
\usepackage{aaai25}  %
\usepackage{times}  %
\usepackage{helvet}  %
\usepackage{courier}  %
\usepackage[hyphens]{url}  %
\usepackage{graphicx} %
\usepackage{natbib}  %
\usepackage{caption} %
\usepackage{algorithm}
\usepackage{algorithmic}
\usepackage{amsfonts}
\usepackage{amssymb}%
\usepackage{pifont}%
\newcommand{\cmark}{\ding{51}}%
\newcommand{\xmark}{\ding{55}}%
\usepackage{newfloat}
\usepackage{listings}
\DeclareCaptionStyle{ruled}{labelfont=normalfont,labelsep=colon,strut=off} %
\floatstyle{ruled}
\newfloat{listing}{tb}{lst}{}
\floatname{listing}{Listing}
\usepackage{xspace}
\usepackage{amsmath}
\newcommand{\name}{\textsc{cyanea}\xspace}
\usepackage{tikz}
\usetikzlibrary{calc}
\usetikzlibrary{positioning}
\usepackage{listings}

\definecolor{codegreen}{rgb}{0,0.6,0}
\definecolor{codegray}{rgb}{0.5,0.5,0.5}
\definecolor{codepurple}{rgb}{0.58,0,0.82}
\definecolor{backcolour}{rgb}{0.95,0.95,0.92}
\definecolor{darkmagenta}{rgb}{0.55,0,0.55}

\lstdefinelanguage{text}{
    basicstyle=\ttfamily\scriptsize,
    sensitive = false,
    keywords={Prompt, Response},
    numbers=none,
    numberstyle=\ttfamily\scriptsize,
    stepnumber=1,
    numbersep=8pt,
    showstringspaces=false,
    breaklines=true,  
}

\lstdefinelanguage{sygus}{
  sensitive = false,
  keywords={declare, fun, synth, check, assert, sat, blocking, define, constraint},
  numbers=left,
  numberstyle=\footnotesize,
  stepnumber=1,
  numbersep=8pt,
  showstringspaces=false,
  breaklines=true,
  comment=[l]{;},
}

\usepackage{pgfplotstable}

\definecolor{blueaccent}{cmyk}{1, 0.3, 0, 0.16}
\definecolor{greenaccent}{cmyk}{1, 0, 0.69, 0.45}
\definecolor{purpleaccent}{cmyk}{0, 0.68, 0.02, 0.49}
\definecolor{orangeaccent}{cmyk}{0, 0.65, 0.79, 0.06}
\definecolor{turquoiseaccent}{cmyk}{0.69, 0, 0.31, 0}

\title{Online Prompt Selection for Program Synthesis}
\author {
    Yixuan Li\textsuperscript{\rm 1},
    Lewis Frampton\textsuperscript{\rm 1},
    Federico Mora\textsuperscript{\rm 2},
    Elizabeth Polgreen\textsuperscript{\rm 1}
}
\affiliations {
    \textsuperscript{\rm 1}University of Edinburgh, UK\\
    \textsuperscript{\rm 2}University of California, Berkeley, USA\\
    \{yixuan.li.cs, elizabeth.polgreen\}@ed.ac.uk
}

\begin{document}

\maketitle

\begin{abstract}
Large Language Models (LLMs) demonstrate impressive capabilities in the domain of program synthesis. This level of performance is not, however, universal across all tasks, all LLMs and all prompting styles. There are many areas where one LLM dominates, one prompting style dominates, or where calling a symbolic solver is a better choice than an LLM. 
A key challenge for the user then, is to identify not only when an LLM is the right choice of solver, and the appropriate LLM to call for a given synthesis task, but also the right way to call it.
A non-expert user who makes the wrong choice, incurs a cost both in terms of results (number of tasks solved, and the time it takes to solve them) and financial cost, if using a closed-source language model via a commercial API. 
We frame this choice as an online learning problem. We use a multi-armed bandit algorithm to select which symbolic solver, or LLM and prompt combination to deploy in order to maximize a given reward function (which may prioritize solving time, number of synthesis tasks solved, or financial cost of solving). We implement an instance of this approach, called \name, and evaluate it on synthesis queries from the literature in ranking function synthesis, from the syntax-guided synthesis competition, and fresh, unseen queries generated from SMT problems. \name solves 37.2\% more queries than the best single solver and achieves results within 4\% of the virtual best solver.
\end{abstract}

\section{Introduction}

Large Language Models (LLMs) are beginning to dominate the discourse around program synthesis and code generation. 
So much so, that one might suppose they are the de facto answer to all code-generation questions. However, this is not the case. There are many synthesis problems in which LLMs still fall far short of the basic enumerative techniques and symbolic solvers~\cite{cav-yixuan,hysynth}.  
In addition, even when an LLM is the best choice, they still hold a significant barrier to entry for the inexperienced user: first, not all LLMs perform uniformly well across all problem sets, and it is often unclear which LLM a user should choose; second, the performance of LLMs is often dependent on careful prompt engineering by expert researchers, with the literature reporting performance gains from many different prompting styles. Finally, compounding the challenge of these choices, calling LLMs is often expensive (in terms of computational cost, or the financial cost of using commercial APIs), and so making the wrong choice for a large set of synthesis tasks is highly undesirable.
This paper addresses these gaps through an  online learning method that, given a synthesis task, will predict whether a symbolic solver or LLM, from a portfolio of LLMs, is most likely to solve the problem, with a corresponding prompting style.

We collate a portfolio of prompting styles and language models, which we combine into LLM-prompt pairs that we refer to as ``solvers''. We formulate the task of ranking the solvers in order of most likely to solve the problem as a multi-armed bandit problem~\cite{multiarmedbandit}. The multi-armed bandit sequentially selects between choices (in our case, solvers) with unknown rewards (in our case, rewards are given for solving problems correctly and fast or with low computational cost). It trades off exploration, i.e., trying new solvers, with exploitation, i.e., using solvers that are known to be good. 
We also present a second variation of this formulation, with multiple layers of bandits. The top multi-armed bandit selects between the symbolic solver and the LLMs, and then the bandits in the lower layer predict the best prompt style for the chosen LLM.

We implement an instance of our approach, \name, and evaluate it on synthesis tasks from the syntax-guided synthesis competition~\cite{sygus-comp}, from the literature on ranking function synthesis~\cite{neuraltermination,DBLP:conf/tacas/GieslRSWY19}, and generated from the SMT competition~\cite{aaaiparsert}. \name solves 37.2\% more synthesis queries than the best single LLM or solver, and gets within 4\% of the virtual best solver.

\section{Background}
\subsection{Program Synthesis}
Program synthesis is the task of automatically generating code or expressions to satisfy some specification. In this paper, we define a program synthesis query $q$ to be a tuple $\langle \tau, f, \phi \rangle$ where $\tau$ is a background theory, $f$ is a function to be synthesized, and $\phi$ is a quantifier-free formula. A solution to the synthesis query is a body for the function $f$ such that the formula $\phi$ is $\tau$-valid, i.e., the formula $\forall x. \phi(f)$ is true where $\phi(f)$ denotes the result of correctly substituting the synthesized body of $f$ into the formula $\phi$, and $x$ is the vector of free variables in $\phi(f)$.

An example program synthesis problem, written in SyGuS-IF~\cite{sygus-if}, is shown in Figure~\ref{fig:example}. Given a candidate solution, we can validate whether this solution is correct or not using a Satisfiability Modulo Theories (SMT) solver, by checking if the formula $\exists x. \neg \phi(f)$ is satisfiable (in which case the candidate $f$ is incorrect) or not.

Program synthesis problems are often accompanied by a context-free grammar $G$, which is a 4-tuple $(V, \Sigma, R, S)$, where $V$ is a finite set of non-terminal symbols. $\Sigma$ with $\Sigma \cap V = \emptyset$ is a set of terminal symbols. $R \subseteq V \times (V \cup \Sigma)^*$ is a finite relation describing the production rules of the grammar. In this work, we do not view the grammar as a syntactic restriction and we provide a grammar that represents the whole space of possible solutions in the logic. 

\begin{figure}[ht!]
    \centering
    \begin{lstlisting}[language=text]
(set-logic LIA)
(synth-fun f ((v0 Int) (v1 Int) (v2 Int)) Int)
(constraint (>= (f v0 v1 v2) v0))
(constraint (>= (f v0 v1 v2) v1))
(constraint (>= (f v0 v1 v2) v2))
(constraint (or (= v0 (f v0 v1 v2)) (or (= v1 (f v0 v1 v2)) (= v2 (f v0 v1 v2)))))
(check-synth)
    \end{lstlisting}
    \caption{A SyGuS specification that asks for a program that synthesizes the maximum of $3$ inputs~\cite{sygus-comp}. We omit the variable declarations for brevity.}
\label{fig:example}
\end{figure}

\subsection{Multi-Armed Bandit Problems}
Multi-armed bandit problems are online-learning problems in which a decision maker iteratively selects from a set of multiple fixed choices. Each choice has an unknown associated reward distribution, and the aim of the decision maker is to maximise the reward achieved over time. Hence, the agent should trade-off exploration (trying new actions to learn more about them) and exploitation (taking actions that are known to have potential for high reward). Here, we frame the problem of choosing LLMs and prompting styles as a MAB problem, where the decision maker selects the LLM and prompt combination to deploy, and the reward is based on successfully solving the synthesis problem. 
We assume that running a solver for a randomly selected synthesis problem is equivalent to sampling from some unknown distribution that we seek to approximate. Contextual MABs extend the problem by giving agents access to a feature vector before each round. This allows us to add information about the characteristics of the synthesis problem
 we are trying to solve in each round.

\section{Overview}

\subsection{Problem Statement}
We hypothesize that program synthesis users will frequently have not just one but a series of synthesis problems to solve. For instance, when synthesizing invariants, one may be synthesising invariants for multiple different loops within the same code base or system under verification. Users may also have different needs when it comes to performance (e.g., fastest, solves most queries, cheapest). 

Given a list of synthesis queries $Q = \{q_1, \ldots q_m\}$, a set of solvers 
$S= \{s_1, \ldots s_n\}$, where each solver is either a symbolic solver, or an LLM paired with a prompting style (and LLM-prompt pair), we wish to use the solvers to generate a list of synthesis functions $f_1, \ldots f_m$ such that each $f_i$ is a valid solution to $q_i$, using as few computational resources as possible. We define computational resources to be both the time spent solving a query, and an estimate of the financial cost of running it (which is based on tokens used for the LLMs, or runtime for the symbolic solver).

\subsection{Approach}
We capture the computational resources we care about as reward functions. Our approach takes in $Q, S$ and a time budget, and cost budget per query, $T$, and $C$ respectively. For each synthesis query, our approach predicts an order of solvers that are most likely to solve the synthesis problem, and the time and cost that each is likely to take. We then distribute the total time and cost budget across the solvers accordingly, and deploy the solvers in sequence until the problem is solved.

After a problem is solved, data is passed back to update the predictors, in the form of any rewards obtained by the solvers called, and the solving time and cost. 

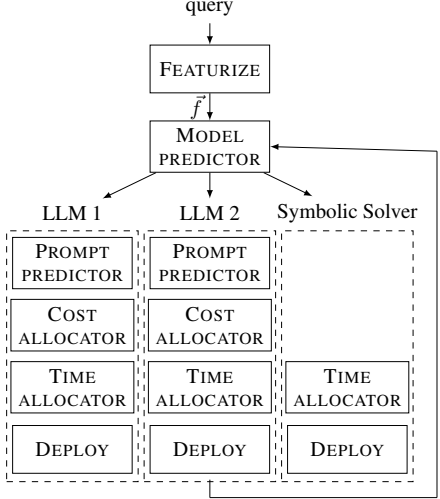
\begin{figure}
\begin{center}
\resizebox{0.7\linewidth}{!}{%
\begin{tikzpicture}[>=latex,x=3cm,y=2cm, mynode/.style={rectangle,draw,minimum height=0.8cm,minimum width=2cm, align=center}]
\def\x{2.3cm}
\def\y{1.3cm}
\node[mynode] at (\x,0.5*\y) (feature) {\textsc{Featurize}};
\node[mynode] at (\x,-0.5*\y) (predict) {\textsc{Model}\\ \textsc{predictor}};

\node[mynode] at (0,-2*\y) (predict1) {\textsc{Prompt}\\ \textsc{predictor}};
\node[mynode] at (0,-2.8*\y) (tokens1) {\textsc{Cost}\\ \textsc{allocator}};
\node[mynode] at (0,-3.6*\y) (time1) {\textsc{Time}\\ \textsc{allocator}};
\node[mynode] at (0,-4.4*\y) (deploy1) {\textsc{Deploy}};

\node[mynode] at (\x,-2*\y) (predict2) {\textsc{Prompt}\\ \textsc{predictor}};
\node[mynode] at (\x,-2.8*\y) (tokens2) {\textsc{Cost}\\ \textsc{allocator}};
\node[mynode] at (\x,-3.6*\y) (time2) {\textsc{Time}\\ \textsc{allocator}};
\node[mynode] at (\x,-4.4*\y) (deploy2) {\textsc{Deploy}};

\node[mynode] at (2*\x,-3.6*\y) (time3) {\textsc{Time}\\ \textsc{allocator}};
\node[mynode] at (2*\x,-4.4*\y) (deploy3) {\textsc{Deploy}};
\node[rectangle,draw, dashed, minimum height=4.2cm,minimum width=2.2cm,align=center] at (0,-3.2*\y) (){};
\node[] at ($(predict1.north)+(0,0.2)$)  (l1) {LLM 1};
\node[rectangle,draw, dashed, minimum height=4.2cm,minimum width=2.2cm,align=center] at (\x,-3.2*\y) (){};
\node[] at ($(predict2.north)+(0,0.2)$)  (l2) {LLM 2};
\node[rectangle,draw, dashed, minimum height=4.2cm,minimum width=2.2cm,align=center] at (2*\x,-3.2*\y) (){};
\node[align=center] at ($(predict2.north)+(\x,0.2)$)  (l3) {Symbolic Solver};

\node[] at ($(feature.north)+(0,0.3)$)  (q) {query};

\path[->] (q) edge[left] node {} (feature);
\path[->] (feature) edge[left] node {$\vec{f}$} (predict);
\path[->] (predict) edge[left] node {} (l1);
\path[->] (predict) edge[left] node {} (l2);
\path[->] (predict) edge[left] node {} (l3);

\path[-] (deploy2.south)+(0,-0.05) edge[above] node {} ($(deploy2.south)+(0,-0.2)$);
\path[-] (deploy2.south)+(0,-0.2) edge[above] node {} ($(deploy3.south)+(0.5,-0.2)$);
\path[-] (deploy3.south)+(0.5,-0.2) edge[above] node {} ($(deploy3.south)+(0.5,2.7)$);
\path[->] (deploy3.south)+(0.5,2.7) edge[above] node {} ($(predict.east)+(0,0)$);

\end{tikzpicture}
}
\end{center}
\caption{Multi-layer Multi-Armed Bandit prediction}
\label{fig:block2-multilayer}
\end{figure}

We break down the task of predicting LLM-prompt pairs in two different ways. In the first, we implement a single multi-armed bandit that predicts between all LLM-prompt pairs at once, show in Figure~\ref{fig:block}. In the second, we implement a multi-stage prediction, where we first predict which solver is most likely to succeed, and then predict which prompt strategy is most likely to succeed if an LLM chosen, shown in Figure~\ref{fig:block2-multilayer}. The components are outlined as follows:

\subsubsection{Featurize:}
The featurize block takes in a synthesis query $q$ in SyGuS-IF and generates a vector of features $\vec{f}$ representing that problem. We use a set of custom-designed features, which are outlined in Section~\ref{sec:mab}.

\subsubsection{Multi-Armed Bandit Solver and Prompt Predictor:}
The multi-armed bandit component of our workflow takes in a feature vector which represents the synthesis problem, and 
 predicts the order in which its library of solvers will perform, according to a cost function. In our implementation, the library of solvers consists of two LLMs, with 6 prompting styles, and an enumerative solver, detailed in section~\ref{sec:prompts}.
 
\begin{figure*}
\begin{center}
\resizebox{0.9\textwidth}{!}{%
\begin{tikzpicture}[>=latex,x=3cm,y=2cm]
\node[rectangle,draw,minimum height=1cm,minimum width=2.2cm] at (0,0) (feature) {\textsc{Featurize}};
\node[rectangle,draw,minimum height=1cm,minimum width=2.2cm] at (1,0) (predict) {\textsc{Predictor}};

\node[rectangle,draw, minimum height=1cm,minimum width=2.2cm,align=center] at (2,0) (tokens) {\textsc{Cost}\\ \textsc{allocator}};
\node[rectangle,draw, minimum height=1cm,minimum width=2.2cm,align=center] at (3.1,0) (time) {\textsc{Time}\\ \textsc{allocator}};
\node[rectangle,draw, minimum height=1cm,minimum width=2.2cm,align=center] at (4.4,0) (deploy) {\textsc{Deploy}\\ \textsc{LLM}};
\node[] at ($(feature.west)+(-0.3,0)$)  (q) {query};
\node[] at ($(deploy.east)+(0.3,0)$)  (r) {result};

\path[->] (feature) edge[above] node {$\vec{f}$} (predict);
\path[->] (predict) edge[above] node {$\{\vec{S}\}$} (tokens);
\path[->] (tokens) edge[above] node {$\{\vec{S},\vec{C}\}$} (time);
\path[->] (time) edge[above] node {$\{\vec{S},\vec{C},\vec{T},\}$} (deploy);

\path[-] (deploy) edge[above] node {} ($(deploy.north)+(0,0.5)$);
\path[-] ($(deploy.north)+(0,0.5)$) edge[above] node {$D = \{r_i,c_i,t_i\}$} node[left,near start,rotate=90,anchor=base,yshift=0.1cm] {} ($(predict.north)+(0,0.5)$);
\path[->] ($(tokens.north)+(0,0.5)$) edge[left] node {$c_i$} ($(tokens.north)+(0,0)$);
\path[->] ($(time.north)+(0,0.5)$) edge[left] node {$t_i$} ($(time.north)+(0,0)$);
\path[->] ($(predict.north)+(0,0.5)$) edge[left] node {$r_i$} ($(predict.north)+(0,0)$);
\path[->] (q) edge[above] node {} (feature.west);
\path[->] (deploy.east) edge[above] node {} (r);

\end{tikzpicture}
}
\end{center}
\caption{Single-layer Multi-Armed Bandit prediction}%
\label{fig:block}
\end{figure*}
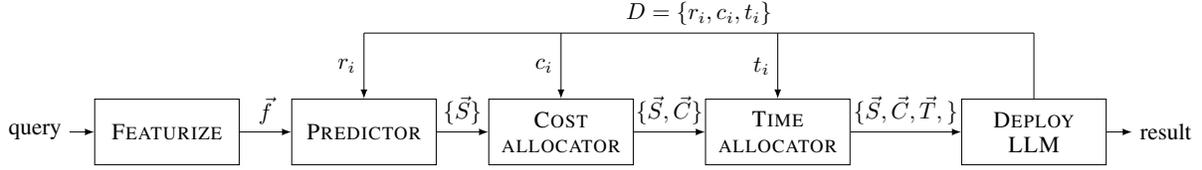

We frame the problem as a contextual multi-armed banded problem, where the actions that the agent is choosing are the solvers. We implement two variations of this: the first, shown in Figure~\ref{fig:block}, uses a single multi-armed bandit agent to rank a set of LLM/prompt combinations and a symbolic solver. The second, layered approach, uses several layered multi-armed bandit agents; one to choose between the base LLMs or symbolic solvers, and a further agent for each LLM, which predicts which prompt to deploy. 

 We give details of the contextual multi-armed bandit algorithms in Section~\ref{sec:mab}. After a solver is deployed, the reward obtained by that solver is passed back to the solver performance predictor, which stores a list of rewards obtained so far by each solver.

\subsubsection{Time and Token Budgeting :}
The final two phases of our pipeline allocate a certain amount of tokens to each LLM, followed by a number of time. Both are described in Section~\ref{sec:mab}.

\subsubsection{Deploy Solvers}
Finally, given a ranked list of solvers, and a time and token budget for each, the deploy phase sequentially calls each solver on the synthesis problem until it either returns an answer or exceeds the time or token budget. If a solver returns the answer, we check if the answer satisfies the specification using an SMT-solver. We also return all information about which solvers successfully solved or did not solve the problem, the reward they obtained, and how long they took to the prediction phases of the pipeline.

\section{Prompting Styles and Solvers}
\label{sec:prompts}
In this section we give an overview of the library of solvers $S$ that our approach is equipped with. First we discuss the prompting styles:

\subsection{Prompting Styles}
We develop a library of prompt templates based on the prompting styles reported to be successful in the literature. We detail the styles here, and illustrate them on our running example. Once we have chosen a style, we give the LLM up to $16$ attempts to produce a correct synthesis result. If an answer produced is incorrect, we report the error information obtained from the SMT-solver used to check correctness back to the LLM:

\subsubsection{Natural language prompts: }
LLMs are primarily trained on natural language inputs, and so we implement a simple syntactic transformation procedure that translates a set of logical constraints into natural language. 

\subsubsection{Few-shot prompting}
Few-shot prompting is prompting whereby the LLM is provided with a number of examples of the task, with satisfying solutions, before asking it to solve a new, similar task. We use $3$ examples, taken from the previously solved synthesis problems.

\subsubsection{Higher resource programming language prompts: } 
Our synthesis queries are in SyGuS-IF, a relatively uncommon language in the training data for LLMs. Thus, we use a prompting style that asks for the solutions in a higher-resource language and then asks for the translation into SyGuS-IF.
We choose Lisp as the higher resource language rather than a more common language like Python because we find that translation from Python to SyGuS-IF is more error-prone than translation from Lisp, which is a fully parenthesized prefix notation similar to SyGuS-IF.
This is a multi-stage prompting approach, the first prompt is shown below:
\vspace{1em}

    \begin{lstlisting}[language=text]
Solve the following function 'solution' with Lisp.
Only return one function, do not use recursion or 
iterations. Do not return any text that isn't code. 
Minimise token use.It's important you keep the 
variables and function names the same as the original 
function. The following is the problem that you are 
meant to solve: 

You need to synthesise: (synth-fun solution ((var1 Int) (var2 Int) Int). The function is called "solution" and takes arguments var1, and var2. These arguments are Int, and Int.
Write only one Lisp-like method "defun solution" that never violates the SMT-LIB constraints.
No built-in functions in code.
Universally quantified variables: var1, and var2. The type of universally quantified variables are Int, and Int.
The function must follow the constraints: 
[constraints]
\end{lstlisting}

\vspace{1em}

The second prompt is as follows:

\begin{lstlisting}[language=text]
Please convert the Lisp function you generated into SMT-LIB format. Follow these guidelines: 
1. Start the function with `(define-fun`.
2. Provide only the function definition, starting with `(define-fun`.
3. Ensure the SMT-LIB function contains exactly one function definition.
4. Avoid using iterations, bitvec, or int notations inside the body.
5. Check the function description in the first message to ensure variable and function names are consistent.
6. Use the assigned values from the Lisp code during translation.
7. Do not introduce any new variables that do not exist in the Lisp function.
8. Pay attention to types. If there are bit-vector terms, ensure they are of the same width.
Rules for SMT-LIB: +, -, *, ite, >, =, <, >=, <=, and, or, not, true, false.
    \end{lstlisting}

When asking for the translation into Lisp, we also provide 3 examples of previous translations.

\subsubsection{Prompting with roles}
Prefixing a prompt with an appropriate role description for the LLM can improve the performance of the LLM~\cite{roles}. If using ``prompting with roles'' we append the sentence \lstinline{You are a good program synthesizer} to the beginning of each prompt. 

\subsubsection{Emotional stimuli}
It has been shown in the literature that adding emotional stimuli to prompts can improve the performance of LLMs~\cite{emotional-prompts}. We append the following emotional stimuli to the prompt.
\begin{lstlisting}[language=text]
You are excited to help, and you are ready to provide 
the best answer possible. You understand that if you 
fail to provide the best answer, your client will be 
extremely upset. Please don't fail me.
\end{lstlisting}

\subsubsection{Matrix of prompts}
In order to reduce the search space of prompts, we choose a fixed combination of prompting styles, shown in Table~\ref{tab:prompt_styles}.

\begin{table}[ht!]
\centering
\begin{tabular}{c|c|c|c|c|c}

 & \rotatebox{90}{Natural Language} & \rotatebox{90}{Higher Resource PL} & \rotatebox{90}{Roles} & \rotatebox{90}{Emotional Stimuli} & \rotatebox{90}{Few-shot} \\ \hline
{Prompt Style 1}  & \cmark                       & \cmark                       & \xmark                    & \xmark                     & \xmark                     \\ \hline
{Prompt Style 2}  & \cmark                       & \cmark                       & \xmark                    & \xmark                     & \cmark                    \\ \hline
{Prompt Style 3}  & \xmark                        & \cmark                       & \xmark                    & \xmark                     & \xmark                     \\ \hline
{Prompt Style 4}  & \xmark                        & \xmark                        & \xmark                    & \xmark                     & \xmark                     \\ \hline
{Prompt Style 5}  & \xmark                        & \cmark                       & \cmark                   & \xmark                     & \xmark                     \\ \hline
{Prompt Style 6}  & \cmark                       & \cmark                       & \cmark                   & \cmark                    & \xmark                     \\ \hline
\end{tabular}
\caption{Prompt styles}
\label{tab:prompt_styles}
\end{table}

\subsection{Enumerative Solver}
The final solver in the library is an enumerative solver, based on CounterExample Guided Inductive Synthesis (CEGIS)~\cite{solar2006combinatorial}, with an $A^*$ based search phase. 
CEGIS alternates between a synthesis phase, which searches for a candidate solution that works for a subset of inputs, and a verification phase, where the candidate is checked against all possible inputs. If the verification fails, a counterexample is passed back to the synthesis phase and appended to the subset of inputs used to guide the search. In our case, the synthesis phase is implemented as an $A^*$ search, similar to that used by Euphony~\cite{euphony}. 

$A^*$ is a graph search algorithm which uses two functions to guide its search: $f$: the sum of the costs on the edges used to reach the current state, and $g$: the estimated sum of the costs on the edges that will be used reach a target state from the current state. In our setting, each state is an expression (a partial or complete program) that can be generated from the grammar for the full logic, the initial state is start symbol of the grammar, the target states are any complete program, and each edge between states $s_i$ and $s_k$ corresponds to a production rule that can transform the partial program at $s_i$ into the partial/complete program at $s_j$. 
The cost on any edge is proportional to the number of possible choices (so the more edges there are leaving from one state, the higher the cost of each edge).

To give some intuition, a partial program that contains few non-terminals and where each non-terminal symbol can only be replaced by production rules that lead immediately to a complete program has a low estimated cost to reach the target. We refer the reader to the detailed descriptions in the related work~\cite{euphony,cav-yixuan} for the full details.
We choose this implementation of CEGIS as the enumerative solver to use because, in our experiments, it excels at finding short solutions that the LLMs often struggle with, without running into the memory issues that often plague bottom-up search methods in synthesis.

\section{Online Solver Selection}
\label{sec:mab}
The aim of the multi-armed bandit is to predict a ranking of which LLM and prompt combinations are most likely to solve the synthesis problem, and obtain the maximum reward while doing so. In our setting, the agent must trade off the exploration of using LLMs and prompts that it has not tried before, vs deploying LLM and prompt combinations that are known to have given high rewards in the past. In fact, we use an extension of the standard MAB problem, and ask the agent to predict a sequence of solvers to deploy rather than a single solver.

\subsection{$k$-Nearest Neighbor}
We choose $k$-Nearest Neighbor as our contextual multi-armed bandit. Other contextual multi-armed bandits are available, but many of the common ones, for instance LinUCB, make assumptions that the performance of solvers is correlated linearly with the feature vector, which is unlikely to be the case in our application.

$k$-NN is a simple supervised learning classifier. In our context, given a synthesis query $q$, it identifies the nearest $k$ previously solved queries to $q$ by calculating cartesian distance between the feature vectors. Each of the $k$ queries $q_1, \ldots q_k$ is labeled with the solver that it was solved by and the reward that was obtained,  $r_1, \ldots r_k$ respectively. The score for a solver $s_i$ is given by the sum of the rewards for all queries solved by $s_i$.
We rank the solvers based on this score (highest score is best). For any solvers that do not appear in this ranking, we randomly shuffle them and append them to the end of the list of solvers. If an LLM-prompt pair solves a query, we add a query with that feature vector to our database of queries with the corresponding reward. 

For the double-layered multi-armed bandit, the first layer contains one $k$-NN multi-armed bandit which selects only between the LLMs, and the second layer contains a $k$-NN multi-armed bandit for each LLM, which selects between prompts. The second layer $k$-NN predictors are independent.

\subsubsection{Reward functions}
Our approach is customizable to different reward functions 
We use three reward functions: the first simply aims to solve the queries as fast as possible, regardless of computational cost; and the second takes computational cost into account. The first reward function is given as follows:
\begin{equation*}
r^{t} = 
\begin{cases} 
0 & \text{if } \text{query } q \text{ is unsolved}, \\
\left(1 - \frac{t}{T}\right)^4 & \text{if } \text{query } q \text{ is solved}
\end{cases}
\end{equation*}
where $t$ is the time taken to solve query $q$, and  $T$ is the total time budget for solving query $q$.

The second reward function aims to prioritize cheaper solving, and so accounts for the number of tokens in the prompt and response. 
\begin{equation*}
r^{c} = 
\begin{cases} 
0 & \text{if } \text{query } q \text{ is unsolved}, \\
\left(1 - \frac{c}{C}\right)^4 & \text{if } \text{query } q \text{ is solved}
\end{cases}
\end{equation*}
where $c$ is a cost estimate proportional to the number of tokens used in solving query $q$ and $C$ is the total cost budget for solving query $q$.
The cost estimate is defined as $c = \text{input tokens} + 3 \times \text{output tokens}$ 
for LLMs, which accounts for the higher cost of output tokens from commercial language model APIs.  The actual cost of deploying an enumerative solver is proportional to the runtime and would be negligible for all queries in comparison to the cost of calling a commercial language model, so we fix the cost for the enumerative solver to be a small constant ($0.4$) for all queries.

The final reward is a simple binary reward, $r^b$, which evaluates to $1$ if a query is solved and $0$ if it is not solved.

\subsubsection{Features}
A key component of the contextual multi-armed bandit is the featurization.  
We propose a feature extraction method to analyze SyGuS queries, capturing key syntactic attributes and query types. The extracted features we use are:
\begin{description}
    \item[Keywords:] Frequencies of specific SMT-LIB keywords (e.g., $+, -, *, \texttt{div},$ etc).
    \item[Query length:] The total number of tokens in the file.
    \item[Constants:] Number of constants of each type.
    \item[Query logic:] e.g., \texttt{BV, LIA, PBE, INV,} etc.
\end{description}

\begin{table*}[ht!]
\begin{center}{%
\begin{tabular}{c|l|l|l|l|l|l|l}%
Solver                         & \% Solved & \# Solved ($r^b$)       & Par-2 Score & $r^c$  & $r^t$  & avg time (s) & avg. cost \\ \hline
Virtual Best                   & 91.8\%    & 1165                    & 23596       & 1106.2 & 1019.2 & 2.4          & 670.3     \\
Single $k$-NN ($r^{c}$)        & 88.3\%    & 1120.6$\pm$7.3          & 37636.3     & 1008.7 & 904.4  & 7.1          & 3122      \\
Single $k$-NN ($r^{t}$)        & 88.2\%    & 1119.1$\pm$8.2          & 37813.7     & 1006   & 905.7  & 7            & 3176.9    \\
Single $k$-NN ($r^{b}$)        & 88.1\%    & 1117.5$\pm$8.3          & 38793       & 995.2  & 888    & 7.6          & 3446.5    \\
Single $k$-NN linear ($r^{t}$) & 87.0\%    & 1104$\pm$0              & 39072       & 1000.9 & 935.2  & 5.5          & 3249.6    \\
Single $k$-NN linear ($r^{c}$) & 87.0\%    & 1104$\pm$0              & 39072       & 1002.1 & 935.2  & 5.5          & 3211.4    \\
Single $k$-NN linear ($r^{b}$) & 87.0\%    & 1104$\pm$0              & 39182.4     & 995.4  & 930.4  & 5.6          & 3394.9    \\
Double $k$-NN ($r^{c}$)        & 84.5\%    & 1071.7$\pm$24           & 53499.3     & 886.9  & 774.7  & 13.1         & 6366.8    \\
Double $k$-NN ($r^{t}$)        & 84.4\%    & 1071.1$\pm$24.7         & 53504.3     & 884.6  & 776.1  & 13           & 6448.3    \\
Double $k$-NN ($r^{b}$)        & 84.3\%    & 1069.8$\pm$24.5         & 53747.4     & 887.8  & 775.2  & 13           & 6262.7    \\
Double $k$-NN linear ($r^{c}$) & 71.8\%    & 910.7$\pm$159.2         & 80038.4     & 803.2  & 732.4  & 9.2          & 4589.1    \\
Double $k$-NN linear ($r^{t}$) & 71.8\%    & 910.7$\pm$159.2         & 80129.5     & 801.9  & 731.8  & 9.3          & 4656.1    \\
Double $k$-NN linear ($r^{b}$) & 71.8\%    & 910.7$\pm$159.2         & 80220.6     & 798.2  & 728.9  & 9.4          & 4809.2    \\
llama-p4                       & 64.3\%    & 816                     & 95251.2     & 752.3  & 664.2  & 5.7          & 2098.2    \\
llama-p3                       & 61.7\%    & 783                     & 106596      & 678.1  & 529.1  & 12           & 3794.1    \\
llama-p5                       & 59.7\%    & 757                     & 112543.8    & 644    & 490.6  & 13.4         & 4275.9    \\
gpt-p4                         & 54.3\%    & 689                     & 118273.7    & 637.1  & 607.4  & 3.3          & 2070.1    \\
enumerator                     & 52.2\%    & 662                     & 122591.6    & 647.3  & 626.7  & 1.8          & 0.4       \\
gpt-p1                         & 51.6\%    & 655                     & 126533.5    & 591    & 531.6  & 5.7          & 2678.1    \\
gpt-p6                         & 50.5\%    & 641                     & 129638.3    & 582.8  & 513.4  & 6.3          & 2508.7    \\
gpt-p5                         & 44.1\%    & 560                     & 145160      & 510.4  & 453.1  & 6            & 2425      \\
gpt-p3                         & 44.0\%    & 558                     & 145101.6    & 502.3  & 464.1  & 5.2          & 2900.5    \\
gpt-p2                         & 39.2\%    & 497                     & 156238.9    & 428.4  & 431.7  & 3.7          & 3755.7    \\
llama-p1                       & 37.2\%    & 472                     & 162184.8    & 436.5  & 380.5  & 5.9          & 2008.6    \\
llama-p2                       & 35.0\%    & 444                     & 168108      & 361.6  & 341.5  & 7            & 5119.5    \\
llama-p6                       & 34.8\%    & 442                     & 167963.6    & 407.3  & 353.8  & 5.8          & 2071.6  
\end{tabular}
\caption{Performance of all instances of \name, and all individual solvers. We report results from \name over 20 runs, with the standard deviation shown for the number of queries solved. The ``Virtual Best'' solver reports the maximum scores we could achieve if we made the perfect choice for each query, using the best reward function for that score (e.g., the score for $r^c$ is reported making choices using $r^c$.}
\label{tab:basic}}
\end{center}
\end{table*}

\subsection{Time and Token Budget Allocation}
The final stage of the dynamic solver selection predicts the time that should be allocated to solver, and the cost. The goal is to allocate a sufficient proportion of our time and cost budget to each solver in the series that we are reasonably confident that it was unlikely to solve the query past this point. That is, for a solver $s_i$, we wish to find a minimum time allocation $t_i$ and cost allocation $c_i$ such that $P(t_i < u_i < T) \leq \delta_1$ and $P(c_i < v_i < C) \leq \delta_2$,  where $u_i$ is the true runtime, $v_i$ is the true cost, and $\delta_1$ and $\delta_2$ are some small error thresholds. $\delta_1$ is the probability that we failed to solve a query because we allocated too little time to solving it, and $\delta_2$ is the probability that we failed to solve a query because we allocated too few tokens to it. 

Let us consider the cost allocation first: we model each prompt-pair's cost per query as an exponential distribution (that is, most queries are solved with a small number of tokens, only a few queries are solved with an excessively large number of tokens). We use maximum likelihood estimation (MLE)~\cite{mle} to estimate the parameters of the underlying exponential distribution, given the costs we have observed so far. Suppose we observe $u_1 \ldots u_n$ costs, which we assume are drawn from an exponential distribution $Exponential(\lambda)$. To find the exponential distribution which fits our observations best, we aim to solve   
$min_{\lambda} n\ln \lambda - \lambda(\Sigma_i u_i)$, where 
$\Sigma_i u_i$ is the sum of all costs observed so far. This gives us the minimizer 
$\lambda^* = \frac{n}{\sum_i u_i}$. 
We can apply the cumulative distribution function and calculate $c_i$ as:
$c_i = \frac{-ln (\delta + e^{-\lambda^*C})}{\lambda^*}.$

To make this contextual, we use only the costs from the $k$ nearest samples, according to the feature vectors. We divide the total token budget $C$ greedily between the solvers, calculation $c_i$ for each solver starting from the beginning of our ranking, and, once we have reached the total budget $C$, all following solvers are allocated zero tokens. If we reach the end of the list and have remaining tokens, they are given to the final solver.

We repeat all of the above for time. It is worth noting that time budget is not independent from cost budget, because if a solver is allocated zero tokens by the cost budget allocator, the time budget allocator will also not allocate it any time.

\section{Related Work}
\paragraph{Automatic prompt generation}
We consider three kinds of existing prompting techniques: \emph{manual}, \emph{continuous}, and \emph{discrete}. All three techniques attempt to improve the performance of LLMs without modifying the LLM itself.

Manual prompting refers to human designed prompt strategies. For example, Chain-of-Thought prompting \cite{chain-of-thought} or Knowledge prompting \cite{liu2021generated} are clever prompting strategies designed by humans that often improve the performance of LLMs across reasoning tasks. We differ from manual prompting in that we take these strategies as inputs and learn to select the best combination of them for a given task. 

Continuous and discrete prompting both refer to automated prompting techniques. Continuous prompting techniques, like Prefix-Tuning~\cite{li2021prefix}, aim to learn domain and task specific vectors that are then used to guide LLMs to better performance. These vectors can usually not be represented by a sequence of tokens, so continues prompting is not often considered interpretable.

In contrast, discrete prompting techniques optimize the text input to the LLM. Some discrete prompting techniques, like PRewrite~\cite{kong2024prewrite}, RLPrompt~\cite{deng2022rlprompt}, TEMPERA~\cite{zhang2022tempera} and GRIPS~\cite{prasad2022grips}, take a starting prompt and search for a new version of the prompt that outperforms the original. To understand the starting prompt, these techniques usually use a second language model in the loop. Other discrete prompting techniques, like AutoPrompt~\cite{shin2020autoprompt}, Liu et al.~\cite{liu2021makes} and Lu et al.~\cite{lu2021fantastically}, generate prompts using templates. AutoPrompt uses LLM gradients to learn task-specific keywords that are then included in the prompt. Liu et al. select examples to include in a few-shot prompt template. In a similar vein, Lu et al. optimize the order of examples. 

Our work is most like the latter category of discrete prompts. We rely on templates and existing prompts to offer inexpensive prompt optimization.

\paragraph{Portfolio solvers}
The term ``portfolio solver'' refers to any algorithm that deploys multiple solvers or solver configurations on a given problem. Most commonly, these are deployed in parallel~\cite{10.1007/978-3-642-02658-4_60}. These techniques share some similarities with prompt ensembling \cite{liu2023pre}, where the answers to multiple queries are combined to improve overall performance. Our work differs in that we aim to improve performance without incurring a significant additional cost. In this sense, our work is most similar to existing approaches that deploy solvers sequentially, like MedleySolver~\cite{DBLP:conf/sat/PimpalkhareMPS21}. For a given input query, MedleySolver predicts a sequence of SMT solvers to deploy based on minimizing Par-2 score (a proxy for time with a penalty for timeouts). 

\paragraph{Other synthesis approaches} Our enumerative solver is based on $A^*$ search~\cite{euphony}. However, it would be possible to incorporate any state-of-the-art SyGuS solver into \name, provided it supports the SyGuS format, including those that combine bottom-up and top-down search~\cite{DBLP:journals/pacmpl/BarkePP20}, or combine learning with enumeration~\cite{DBLP:conf/iclr/OdenaSBSSD21}.

\section{Evaluation}

We implement an instance of our approach, called \name, using two LLMs: GPT(gpt-3.5-turbo-0125) and Llama(Meta-Llama-3-70B) and the enumerative solver described previously. We set a total timeout of $T = 100$ seconds and a total cost budget of $C = 100,000$. For all $k$-NN predictors, we set $k = 15$. We conducted a parameter sweep of $k$ and found that values between $10$ and $15$ produce comparable results. We use cvc5~\cite{cvc5} as the SMT solver for validating the correctness of candidate solutions. We compare our approach to the base solvers and to a ``virtual best'' solver result, which is calculated by choosing the solver known to give the highest reward for each query.
To evaluate the utility of the time and budget allocations, we compare to a linear distribution of the time and cost budget (where we simply divide the total budget equally between all solvers), termed ``linear'' in the results.  

\subsubsection{Synthesis Queries and Scoring}
We evaluate our approach on synthesis queries from the Syntax-Guided Synthesis competition~\cite{sygus-comp}, ranking function synthesis~\cite{neuraltermination}, and automatically generated from the SMT competition~\cite{aaaiparsert}. The synthesis queries cover a broad range of use-cases of synthesis, from code generation and programming-by-example, to ranking function and invariant synthesis. The total number of queries is $1269$.

 We report the total reward achieved, calculated using the reward functions used for the prediction. We also report the score according to the Par-2 score used by the SAT competition~\cite{WCDHNR19}. This is calculated over $n$ queries: 
 $$\sum_{j=1}^{j=n} \begin{cases}
        t_i & \text{if $q_i$ is solved},\\
        2*T & \text{otherwise}
    \end{cases}$$
    where $t_i$ is the runtime for solving query $q_i$, and $T$ is the total time budget per query.

\begin{figure}
    \centering
    \includegraphics[width=0.9\linewidth]{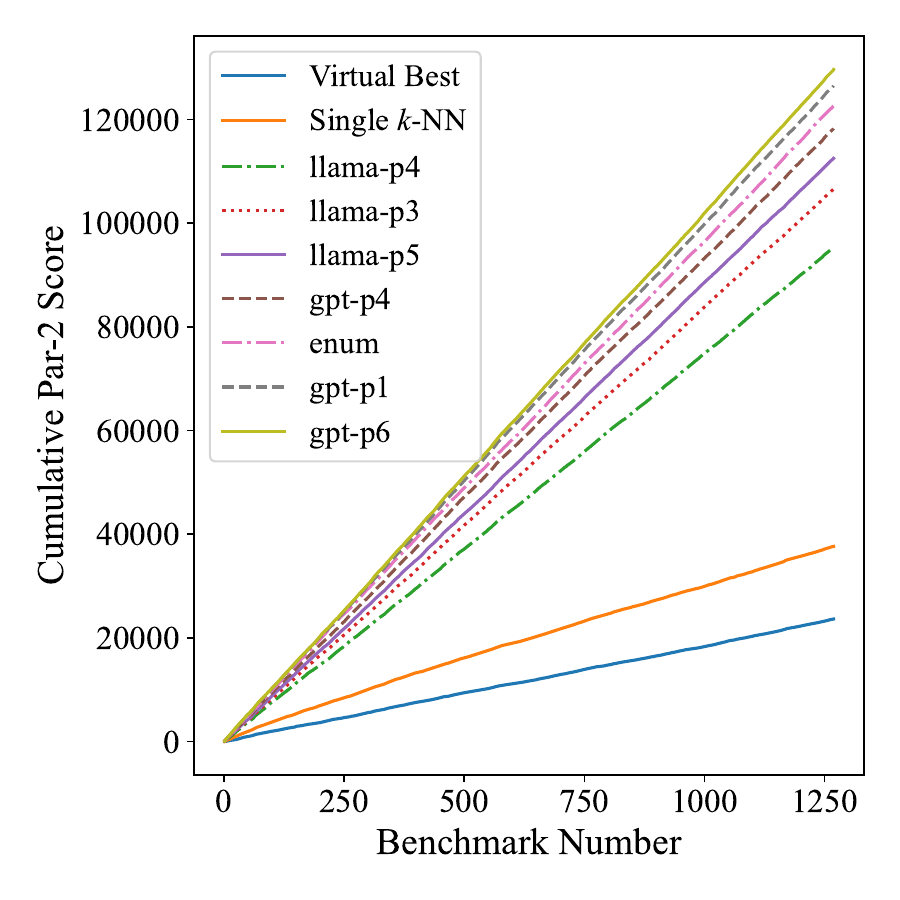}
    \caption{Cumulative Par-2 Score plotted against query number. Lower is better.}
    \label{fig:par2plot}
\end{figure}

\subsubsection{Analysis of Results}
The performance of the LLM-prompt pairs and enumerative solver is shown in Table~\ref{tab:basic}. The best-performing single solver solves $64.3\%$ of the queries. On the other hand, the virtual best solver solves $91.8\%$ of queries. 
The best-performing instance of \name solves $88.3\%$ of the queries, achieving a score of $96.1\%$ of the virtual best solver, and a Par-2 score $<40\%$ of the score of the best single solver (lower is better for Par-2 score), shown in Figure~\ref{fig:par2plot}.

Since the order the selectors see the queries will affect the learning, we report the average scores of $20$ runs, with the queries randomly shuffled in each run. When comparing the single to multi-layer approaches, we note that the multi-layer approach has a far greater standard deviation in the number of queries solved. We hypothesize that this 
is because the 
selectors in the lower layers see far less data than in the single-layer $k$-NN, making them more sensitive to this ordering.

When evaluating the reward functions used by single $k$-NN,  unsurprisingly the best Par-2 score (which is based on solving time) is achieved using $r^t$, the lowest cost per query is achieved using $r^c$, and the highest number of queries solved is achieved using $r^b$. This is not true for double $k$-NN, again we believe because the data is too sparse. 

The results for \name using the linear distribution of time and token budgets demonstrate that the token budget allocator is having an impact on both Par-2 score and average solving cost, although this effect is not as large as it would be if the total time budget and cost budget were tighter, as in many cases \name can run all the solvers on a single query within the given budget.  Whilst the double $k$-NN with the linear distribution of budgets does have lower average costs per query, we hypothesize that this is because it fails to solve many of the queries that require a higher cost to solve.

Overall, the predictions that the best-performing instance of \name is making are close to those of the virtual best solver and, when the prediction is not perfect, the time/budget allocation allows \name to correct the mistake.

\subsection{Conclusions}
We have presented an approach for online solver and prompt selection for program synthesis problems; \name demonstrates the effectiveness of this, achieving a Par-2 score that is more than twice as good as the best single solver. 

\subsubsection{Acknowledgments } This work was supported in part by a Royal Academy of Engineering Research Fellowship and an Amazon Research Award.

\bibliography{Li}

\end{document}